\documentclass[sigconf]{acmart}
\usepackage{ulem}
\usepackage{listings}
\AtBeginDocument{%
  }

\setcopyright{acmlicensed}
\copyrightyear{2018}
\acmYear{2018}
\acmDOI{XXXXXXX.XXXXXXX}

\acmConference[Under Review]{}{}{Perth, WA}
\acmISBN{xxx-xxx-xxx}

\begin{document}

\title{Docs2KG: Unified Knowledge Graph Construction from Heterogeneous Documents Assisted by Large Language Models}

\author{Qiang Sun}
\email{pascal.sun@research.uwa.edu.au}
\orcid{0000-0002-4445-0025}
\affiliation{%
  \institution{The University of Western Australia}
  \city{Perth}
  \state{WA}
  \country{Australia}
}

\author{Yuanyi Luo}
\email{luoyy@stu.hit.edu.cn}
\affiliation{%
  \institution{Harbin Institute of Technology}
  \city{Harbin}
  \country{China}}

\author{Wenxiao Zhang}
\email{wenxiao.zhang@research.uwa.edu.au}
\orcid{0009-0000-5196-8562}
\affiliation{%
  \institution{The University of Western Australia}
  \city{Perth}
  \state{WA}
  \country{Australia}
}

\author{Sirui Li}
\email{sirui.li@uwa.edu.au}
\orcid{0000-0002-2504-3790}
\affiliation{%
  \institution{The University of Western Australia}
  \city{Perth}
  \state{WA}
  \country{Australia}
}

\author{Jichunyang Li}
\email{jichunyang.li@uwa.edu.au}
\orcid{0009-0008-3075-3739}
\affiliation{%
 \institution{The University of Western Australia}
  \city{Perth}
  \state{WA}
  \country{Australia}
}

\author{Kai Niu}
\email{kai.niu@research.uwa.edu.au}
\orcid{0009-0009-3357-6130}
\affiliation{%
  \institution{The University of Western Australia}
  \city{Perth}
  \state{WA}
  \country{Australia}
}

\author{Xiangrui Kong}
\email{xiangrui.kong@research.uwa.edu.au}
\orcid{0000-0001-5066-1294}
\affiliation{%
  \institution{The University of Western Australia}
  \city{Perth}
  \state{WA}
  \country{Australia}
}

\author{Wei Liu}
\email{wei.liu@uwa.edu.au}
\orcid{0000-0002-7409-0948}
\affiliation{%
 \institution{The University of Western Australia}
  \city{Perth}
  \state{WA}
  \country{Australia}
}

\renewcommand{\shortauthors}{Sun et al.}

\begin{abstract}
Even for a conservative estimate, 80\% of enterprise data reside in unstructured files, stored in data lakes that accommodate heterogeneous formats. Classical search engines can no longer meet information seeking needs, especially when the task is to browse and explore for insight formulation. In other words, there are no obvious search keywords to use. Knowledge graphs, due to their natural visual appeals that reduce the human cognitive load, become the winning candidate for heterogeneous data integration and knowledge representation.  
% Given the nature of unstructured and heterogeneous data, information extraction and knowledge representation pose significant challenges. 
In this paper, we introduce Docs2KG, a novel framework designed to extract multimodal information from diverse and heterogeneous unstructured documents, including emails, web pages, PDF files, and Excel files. Dynamically generates a unified knowledge graph that represents the extracted key information, Docs2KG enables efficient querying and exploration of document data lakes. Unlike existing approaches that focus on domain-specific data sources or pre-designed schemas, Docs2KG offers a flexible and extensible solution that can adapt to various document structures and content types. The proposed framework unifies data processing supporting a multitude of downstream tasks with improved domain interpretability. Docs2KG is publicly accessible at \url{https://docs2kg.ai4wa.com}, and a demonstration video is available at \url{https://docs2kg.ai4wa.com/Video}.

%Despite their superb performance and general applicability, LLMs still struggle with faithful information seeking for domain-specific natural language processing tasks. The two possible solutions, fine-tuning and Retrieval Augmented Generation (RAG), both require some domain-specific data to enable LLMs' domain transferability. However, domain data are often available in diverse and unstructured formats, such as web pages, emails, and PDF files, which poses significant preprocessing challenges for effective LLMs consumption.

%In this paper, we introduce Docs2KG, a framework designed to transform unstructured data into a unified knowledge graph. Docs2KG enables the integration of diverse data formats, facilitating efficient preprocessing. The generated KG can be used for space-efficient fine-tuning and knowledge-grounded RAG, which improves the domain applicability of LLMs. The proposed framework not only simplifies data processing but also improves the interpretability and utility of LLMs across diverse domains. 
\end{abstract}


\keywords{Unstructured Data, Heterogeneous Data, Knowledge Graph}
% TODO: add a overall architecture
%   \includegraphics[width=\textwidth]{imgs/Docs2KG_Design.pdf}
%   \caption{Architecture Design for \textbf{Docs2KG}}
%   \Description{End to end workflow about how Docs2KG is designed.}


\maketitle

\section{Introduction}
The most valuable enterprise knowledge reside in unstructured documents of heterogeneous formats, taking up at least 80\% of the corporate data lakes. It is crucial to extract meaningful information~\cite{DBLP:conf/coling/LuoDLPH22} by integrating these data, while maintaining references to the origin for Retrieval Augmented Generation (RAG)~\cite{lewis2020retrieval} to reduce hallucination.  %Recognising the layout and components of heterogeneous unstructured documents and extracting meaningful information from them are crucial tasks~\cite{DBLP:conf/coling/LuoDLPH22}. 
Taking the healthcare industry as an example, patient records often exist in various formats such as handwritten clinical notes, discharge letters, email communication between clinicians, and medical images. Without data integration, it is impossible to provide a consolidated assessment. Many existing works~\cite{DBLP:conf/coling/LuoDLPH22, DBLP:conf/coling/LiXCHWLZ20} are designed to target a single data source, such as scanned documents or PDF files. However, in real-world applications, particularly within domain-specific knowledge areas, data are heterogeneous, unstructured, and diverse~\cite{maree2015addressing}. To perform document-wide semantic parsing and layout analysis from heterogeneous unstructured documents, we face three key challenges:
%However, real-world applications, particularly domain knowledge, often involve multimodal data, such as images and tables, which are hidden in emails, web pages, PDF files and Excel files. For example, clinicians cannot fully understand the extent of internal injuries in a trauma patient without the detailed images provided by a CT scan. It is then crucial to parse diverse unstructured documents and analyse the document layout. To implement it, we face two key challenges:
\begin{itemize}
% \item The dynamic analysis for multimodal data (incl. tables, texts, images, and figures) that are hidden in diverse and heterogeneous unstructured data sources.
% \item The physical analysis involves detecting the document structure and layout.
% \item The spatial analysis involves organising the detected layout components based on the document title and header.
% \item The semantic analysis involves considering sentence-level meanings and context.
\item The extraction of multimodal data (incl. tables, texts, images, and figures) from a diverse range of formats.
\item Integrating modality-specific information extraction models into one unified framework.
\item Meaningful representation of data semantic with references to the source. 
\end{itemize}
\begin{figure*}[htbp]
  \includegraphics[width=0.95\textwidth]{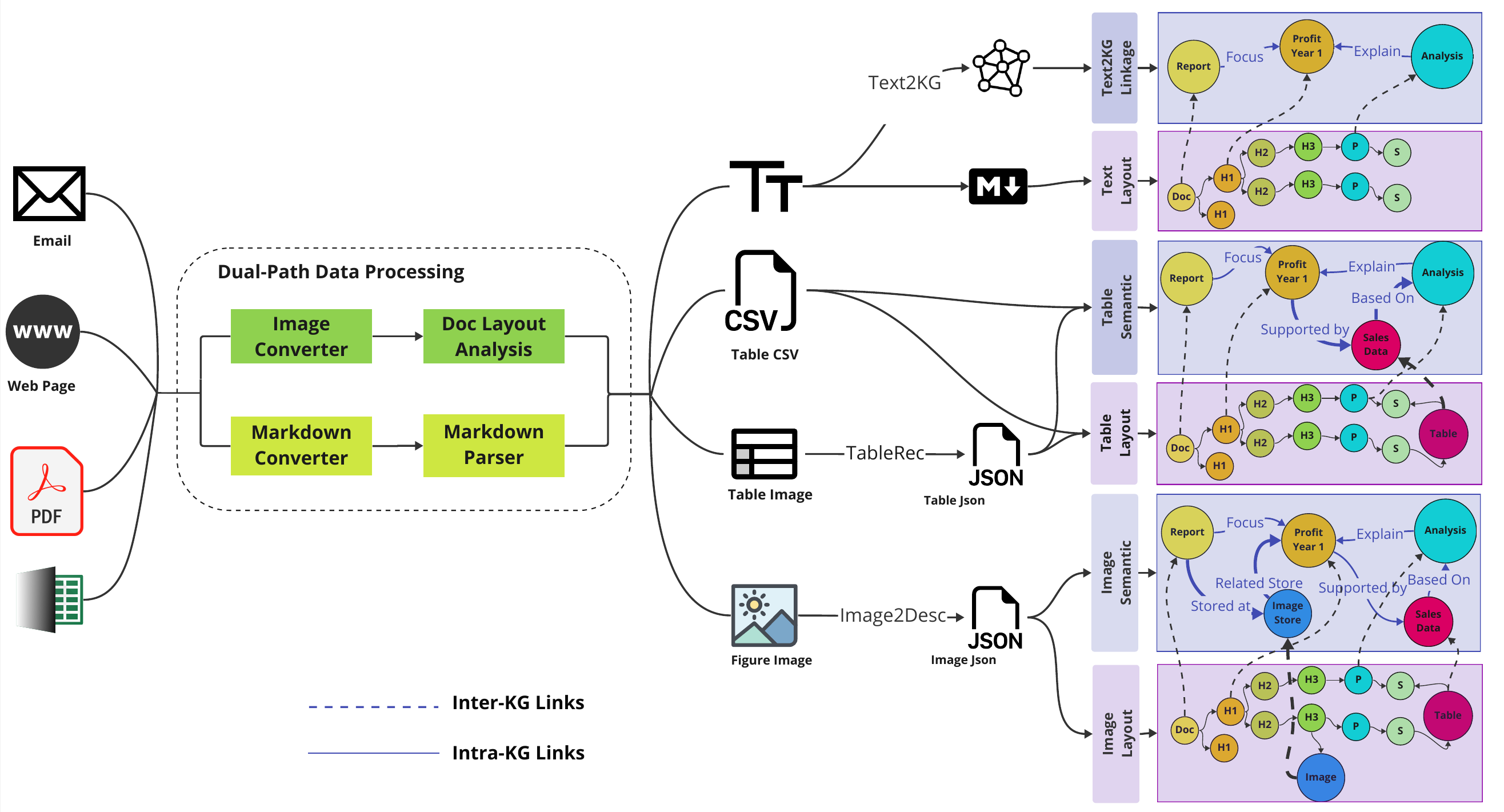}
  \caption{Architecture Design for \textbf{Docs2KG}}
  \Description{End to end workflow about how Docs2KG is designed.}
  \label{fig:teaser}
\end{figure*}

In this research, we propose using Knowledge Graphs as a unified representation to allow dynamic integration of entities extracted from each modality, including layout entities to maintain references to the source. The end goal of knowledge graph construction is faciliated through our proposed \textbf{Docs2KG} system to address the above challenges.% by extracting multimodal information from diverse and heterogeneous data sources. 
The data formats that Docs2KG can handle include emails, web pages, PDF files, and Excel files. The extracted multimodal information, merged as a unified KG, allows for dynamic and automatic update based on document structure and content, which can be modified and extended to allow human-in-the-loop. It enables researchers and domain experts to pose structural and semantic queries such as \textit{``Show me all documents and their components related to events that occurred in the years 2011 and 2021.``}. This capability can dramatically reduce the time, effort, and resources required navigating through large collections of unstructured documents. Moreover, Docs2KG unified document processing through \textit{a dual path strategy} which effectively combined deep learning computer vision based document layout analysis with mark-down structured document parsing to maximise its document type coverage. The KG generated by Docs2KG can be used to facilitate many real-world applications, such as reducing the risk of outdated knowledge and hallucination of language language models to achieve knowledge-grounded retrieval augmented generation.  %improves the interpretability and utility of models across diverse domains.
\section{Related Work}
There have been several efforts to construct KGs to facilitate the discovery of relevant information within specific fields. Most of these efforts~\cite{DBLP:conf/aaai/Kertkeidkachorn17, DBLP:conf/semweb/RossanezR19, eitan2021connected} have focused on extracting information from text. For example, Connected Papers~\cite{eitan2021connected} is a tool designed to help researchers and academics to find and explore relevant academic papers. It creates a citation network of papers for a given search paper, allowing users to see connections and discover influential works in their field. This visualisation aids in the literature search in a broader context assisting in finding seminal works and new directions worth investigation. Another example is the work by \citet{kannan2020multimodal}, who built a multimodal KG that extracts text, diagrams, and source code from scientific literature in the field of Deep Learning. 

Our framework, Docs2KG, differs from these approaches by specifically targeting at heterogeneous unstructured documents rather than just scientific publications. While their schema is predesigned for specific domains, such as deep learning architectures, ours is dynamic and automatically generated based on the document structure. Additionally, Docs2KG can be modified and extended as needed, making it more adaptable to various types of unstructured data.

\section{Docs2KG Framework}
\label{section:docs2kg}
The architecture of Docs2KG is shown in Figure~\ref{fig:teaser}, which is designed to take asinput a set of heterogeneous and unstructured documents, including emails, web pages, PDF files and Excel files. Docs2KG involves two main stages: dual-path data processing and multimodal unified KG construction. The \textit{dual-path data processing} stage segments the input documents into textual content, images, and tables. The \textit{multimodal unified KG construction stage} integrates the processed information with structural and semantic relationships.

%Specially:
%\begin{itemize}
%    \item Textual content is extracted and converted to Markdown format to facilitate spaitial layout anaysis.
%    \item Figures and tables are extracted and converted into JSON format for structured representation to facilitate semantic and spatial analysis.
%\end{itemize}
After alignment, the resulting multimodal KG is stored in a Neo4j\footnote{\url{https://neo4j.com/}} graph database, allowing storage of the extracted information a triple store for efficient querying and intuitive visualisation. All code and documentation are available online\footnote{\url{https://docs2kg.ai4wa.com/}}. The code is designed to be modualised, other graph databases can be used to replace Neo4j for graph data storage and retrieval. The following sections detail the two key stages of Docs2KG.

\subsection{Dual-Path Data Processing}
\begin{figure*}[!htbp]
  \includegraphics[scale=0.25]{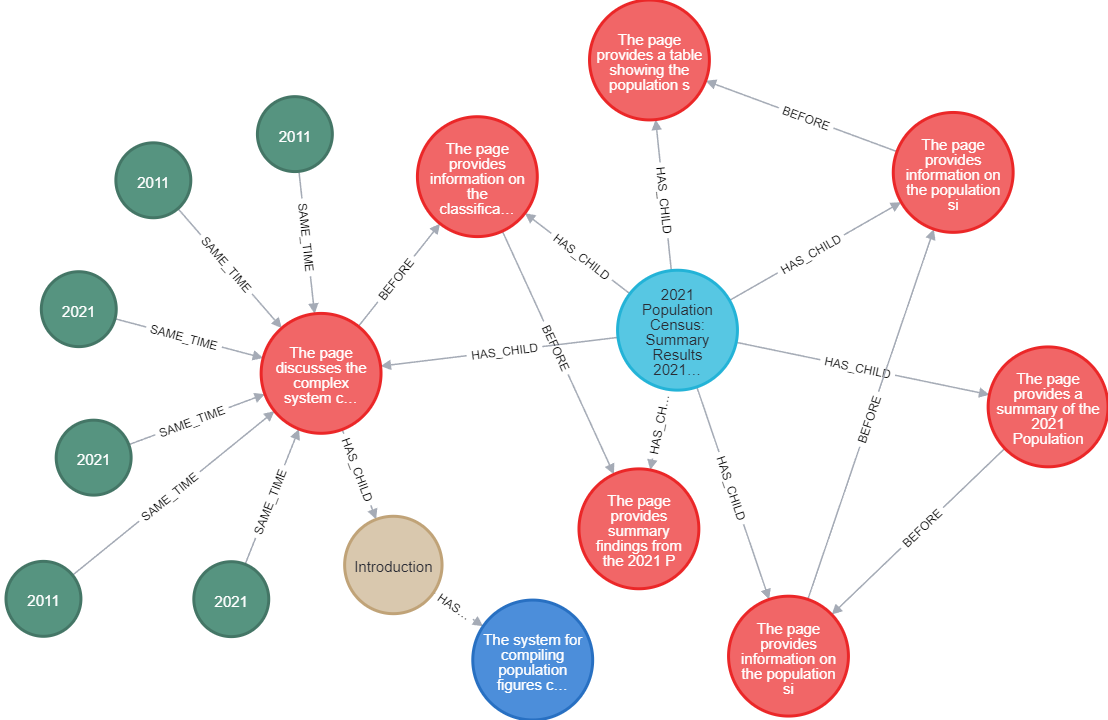}
  \caption{A demo graph of query ``Show me all documents and their components related to events that occurred in the years 2011 and 2021.'' by combining a PDF file and an Excel file. The PDF file contains information about the population size and structure of Hong Kong from 2011 to 2021. The Excel file contain records of the population census from 2021 to 2023. (Cyan indicates the PDF document; Green is for Excel file; Red for PDF page; Khaki for header; ocean blue for paragraph)}
  \label{fig:demo_query_graph}
\end{figure*}
In Figure~\ref{fig:teaser}, we categorise the input documents into two types based on the easy of extracting their layout information. For example, web pages (HTML) are organised using a tree structure, enabling straightforward conversion to Markdown or JSON. In contrast, PDF files and Excel files with extensive descriptive information pose significant challenges for layout detection and transformation into semi-structured format. To address the above challenges, we propose a dual path document processing strategy. The\textit{ Image Converter} path is a generic approach that uses deep learning models trained for document layout analysis; the \textit{Markdown Converter} path is to convert documents to markdown format and use an XML/HTML query language such as XPath. All four types of documents can be converted into images and take advantage of the document layout analysis to segment into texts, images, and tables with bounding boxes. We will not provide details on how these are achieved; please refer to our publications on PDF form data analysis~\cite{Wu:2022}. For markdown document parsing, we have developed four independent parsers to process different document types:
\begin{itemize}

\item \textbf{PDF parsing:} Based on the meta information provided by the PDF file, we can determine whether to feed it to the \textbf{Markdown Converter} or \textbf{Image Converter}. For scanned PDF files, the only path is through trained document layout analysis models, while generated PDF files can be parsed or segmented to extract images, tables, texts with bounding box information.

\item \textbf{Web page parsing:} We use a popular Python library, \\ \texttt{BeautifulSoup}~\cite{Hajba2018}, for efficient HTML parsing. Texts are extracted using \texttt{markdownify} \cite{markdownify}. Images are identified via the \texttt{<img>} tag, tables via the \texttt{<table>} tag. The original document tree structure of the HTML page is retained as alayout knowledge graph.

\item \textbf{Excel parsing:} Using the Python library \texttt{pandas}, Excel files are loaded and data are extracted from each worksheet. The extracted data is then converted into images via \textit{imgkit}, and then go through the \textbf{Image Convertor} path. For complex structured Excel worksheets, they can converted to PDF files first, to follow the PDF processing pipeline. 

\item \textbf{Email parsing:} We assume emails are in .eml format. The Python library \texttt{email} is then used to segment messages into plain text, HTML, and attachments. Text and HTML sections of the emails can then be processed similarly to web pages, while attachments are handled by appropriate tools based on their formats, such as PDF or Excel parsers.

\end{itemize}

By combining parsers and document segmentation models, Doc2KG can parse different heterogeneous and unstructured documents for subsequent integration into a unified KG. The modualised approach we are taking allow for flexible configuration and combination of the processing modules to optimise computation resource usage. 

\subsection{Multimodal Unified Knowledge Graph Construction}
After the first stage, our proposed Docs2KG unifies the parsed information into a multimodal KG containing structural (hierarchical and spatial) and semantic information. 

We categorise relationships of our multimodal KG into two primary types: intra-modal relationship and inter-modal relationship.

\textbf{Intra-modal relationships construction:} Intra-modal relationships include structural relationships at the title level and paragraph level, and semantic relationships at the sentence level. The intra-modal relationships can be expressed as:
\begin{equation}
    G^{(\alpha, \beta)}=(h_\alpha, r, t_\beta), \alpha\neq\beta\in \{T, P, S\}
\end{equation}
where the $G$ represents a smallest unit sub-graph in our multimodal KG. $\alpha$ and $\beta$ represent different modalities from text source, containing text ($T$), paragraph ($P$), and sentence ($S$). The notation $(h_\alpha, r, t_\beta)$ denotes the construction method between two nodes, where $h_\alpha$ (the head entity) points towards $t_\beta$ (the tail entity). $r$ denotes the relationship, expressed with structural or semantic information:
\begin{itemize}
    \item \textbf{Structural relationships:} `has-child', `before' and `after'.
\item \textbf{Semantic relationships:} `same time', `focus', `supported by', `explain'.
\end{itemize}
\textbf{Inter-modal relationships construction:} We use semantic relationships to express the relationships between different modalities. It is because the intra-modal hierarchical and spacial relationships already provide a clear relationship direction. The inter-modal relationships can be expressed as:
\begin{equation}
    G^{(S, M)}=(h_S, r, t_M), M\in \{Table, Figure\}
\end{equation}
where $G$ represents a smallest unit sub-graph. $S$ denotes sentences, such as table captions. $M$ denotes tables and figures. $r$ is the semantic relationship between them: `explain' and `same-time'.
\section{Demonstration}
In our demonstration, we first focus on how our multimodal KG can be utilised to perform data-driven analysis through a graph querying demo. Subsequently, we demonstrate how the KG can support one of the most important applications of large language models, RAG. In our RAG demo, nodes and relationships are embedded and subjected to a similarity search to identify anchor nodes. These nodes are then expanded via multi-hop queries to retrieve relevant information, thereby augmenting the prompt to respond to the query.
\subsection{Knowledge Graph Query}
We selected one PDF file and one Excel file for the demo. The PDF file contains information about the population size and structure of Hong Kong from 2011 to 2021. The Excel file contain records of the population census from 2021 to 2023, including mid-year population data categorised by age group and sex.

\begin{figure}[h]
  \includegraphics[scale=0.4]{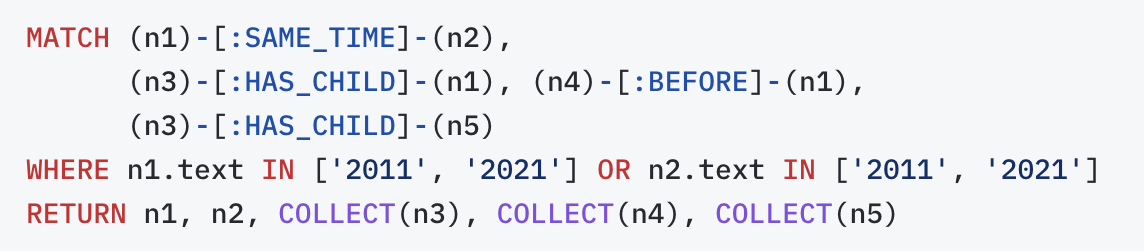}
  \caption{The Cypher Query to answer ``Show me all documents and their components related to events that occurred in the years 2011 and 2021."}
  \label{fig:demo_query}
\end{figure}

Meaningful insights cannot be derived from either the Excel file or the PDF file alone. We parsed and integrated the PDF file and the Excel file through Docs2KG. The data were extracted into figures, tables, and text, and merged into a single KG. To extract relevant information, we used the query shown in Figure~\ref{fig:demo_query}. The returned graph is in Figure~\ref{fig:demo_query_graph} where green bubbles and red bubbles represent the information extracted from Excel and PDF files, respectively. Based on the visualisation, we can observe that the introduction section (the Khaki coloured node) of the PDF document references several events occurring in both 2011 and 2021. For more information about this demo, please refer to our demo video \url{https://docs2kg.ai4wa.com/Video/}.

\subsection{Semantic and Structural Proximity-Based Information Retrieval}
To enhance the performance of large language models, the RAG approach suggests integrating more relevant information directly into the prompt. In the context of our multimodal knowledge graph, `relevance' refers to the proximity of nodes, which can be either semantic or structural. Specifically, relevant nodes are those that can be reached within a limited number of hops in the knowledge graph.
\begin{figure}[h]
  \includegraphics[scale=0.18]{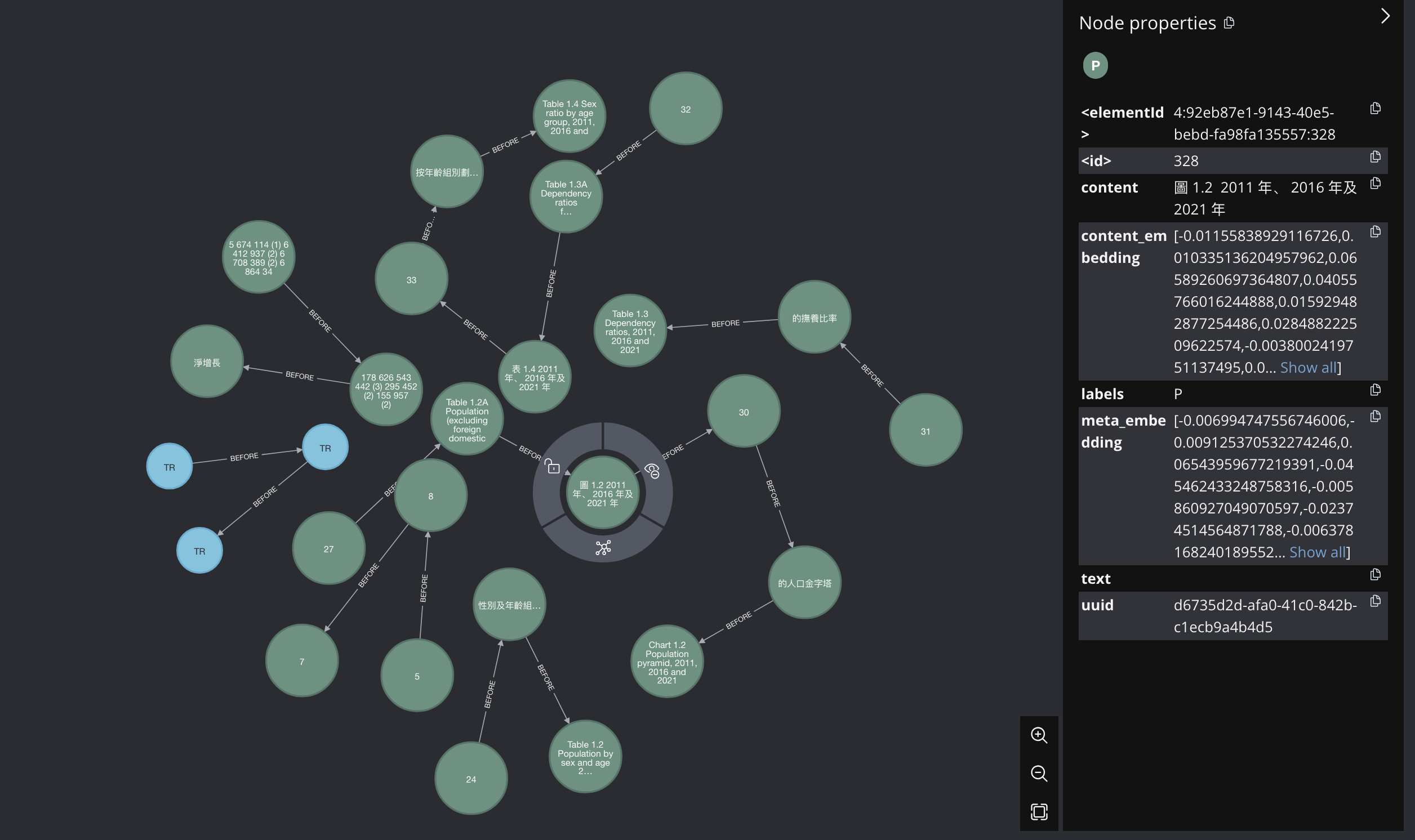}
  \caption{Retrieved relevant semantic and structural nodes for query ``I want to know all the population information from 2011 to 2021" by combining the same files referenced in Section 4.1. (Green indicates \texttt{<p>} tag; Blue for \texttt{<tr>} tag.)}
  \label{fig:rag_demo}
\end{figure}

Based on this, consider the same query in above demonstration: ``I want to know all the population information from 2011 to 2021". Initially, all nodes within the knowledge graph are embedded using an embedding model. The same model is used to embed the query. The query embedding is then utilized to retrieve relevant text chunks, figures, and tables through semantic similarity search. The top-k semantically relevant nodes will be selected as anchor nodes to retrieve the n-hop semantic and structural relevant nodes, there by augmenting the prompts as shown in Figure \ref{fig:demo_query}. We can see the tables regarding the population information from 2011 to 2021 are retrieved.
For additional details regarding this demonstration, please refer to our demo video at \url{https://docs2kg.ai4wa.com/Video/} or our codes.

\section{Conclusion}
In this paper, we have addressed the limitations of existing multimodal KG construction methods by proposing an open-source framework, Docs2KG. Unlike previous approaches that either focus solely on images or rely on an existing KG to link images, our framework considers more realistic scenarios across all domains. Docs2KG effectively handles the diversity and heterogeneity of raw data in various unstructured formats, such as web pages, emails, PDF files, and Excel files. By integrating these diverse data sources into a unified KG and incorporating both semantic and structural information, Docs2KG enables a more comprehensive and accurate representation of knowledge. This facilitates a wide range of real-world applications, improving the utility and robustness of KGs in diverse domains.
\bibliographystyle{ACM-Reference-Format.bst}
\bibliography{refer.bib}

\end{document}